\pdfoutput=1

\documentclass[11pt]{article}

\usepackage[]{emnlp2021}

\usepackage{animate}
\usepackage{fontawesome}

\usepackage{times}
\usepackage{latexsym}

\usepackage[T1]{fontenc}

\usepackage[utf8]{inputenc}

\usepackage{microtype}

\usepackage{graphicx}
\usepackage{booktabs}
\usepackage{xcolor}
\usepackage{xspace}
\usepackage{graphicx}
\usepackage{subfig}
\usepackage{multirow}
\usepackage{array}
\usepackage{verbatim}
\usepackage{mdwlist}
\usepackage{listings}
\usepackage{supertabular}
\usepackage{graphicx}
\usepackage{animate}
\usepackage{amssymb}
\usepackage{makecell}
\usepackage{grffile}

\newcommand{\myparagraph}[1]{\noindent \textbf{#1}}

\newcommand{\pepe}{{{\textsc{Pepe}}}\xspace}

\newcommand{\Sref}[1]{\S\ref{#1}}

\newcommand{\fref}[1]{Figure~\ref{#1}}
\newcommand{\tref}[1]{Table \ref{#1}}

\newcommand{\vgif}[1]{
\begin{minipage}{0.15\textwidth}
\href{http://giphy.com/gifs/#1}{
\includegraphics[width=\textwidth]{gifs/#1.jpg} }
\end{minipage}
}

\newcommand{\myvgif}[1]{
\begin{minipage}{0.12\textwidth}
\href{http://giphy.com/gifs/#1}{
\includegraphics[width=\textwidth]{gifs/#1.jpg} }
\end{minipage}
}

\title{An animated picture says at least a thousand words: Selecting Gif-based Replies in Multimodal Dialog}

 \author{
    Xingyao Wang \\
    University of Michigan \\
    \texttt{xingyaow@umich.edu}\\ 
    \And
    David Jurgens \\
    University of Michigan \\
    \texttt{jurgens@umich.edu}\\}
   
\date{}

\begin{document}
\maketitle
\begin{abstract}
Online conversations include more than just text. Increasingly, image-based responses such as memes and animated gifs serve as culturally recognized and often humorous responses in conversation. However, while NLP has broadened to multimodal models, conversational dialog systems have largely focused only on generating text replies. Here, we introduce a new dataset of 1.56M text-gif conversation turns and introduce a new multimodal conversational model \textsc{Pepe the King Prawn} for selecting gif-based replies. We demonstrate that our model produces relevant and high-quality gif responses and, in a large randomized control trial of multiple models replying to real users, we show that our model replies with gifs that are significantly better received by the community.
\end{abstract}

\section{Introduction}

Conversations are central to many online social platforms. While most conversations are text-based, computer mediated dialog also affords alternative forms of communication, such as emoji or stickers like bitmoji, that allow users to express themselves \cite{tang2019emoticon,konrad2020sticker}. Increasingly, these visual forms of communication have become common in social media \cite{bourlai2014multimodal,highfield2016instagrammatics}, with a notable use of the reaction gif \cite{bakhshi2016fast,miltner2017never}.  These gifs are short video sequences that depict a particular scene and sometimes contain text that acts as a meta-commentary \cite{eppink2014brief}. As a result, conversations become \textit{multimodal} where individuals reply to one another using combinations of text and gifs (\fref{fig:intro-example}). While conversational AI systems have been developed in a purely text-based setting, such systems do not capture the full multimodal behavior seen online. Here, we study multimodal conversation by introducing new dialog models for selecting gif replies in conversation.

\begin{figure}[t]
    \texttt{PizzaMagic}: Ahhhhh!!! The EMNLP deadline is in 24 hours!! \\
    \makecell{ \hspace{-35mm} $\llcorner$ \texttt{CasualModel}: \\
    \begin{frame}{}
        \animategraphics[autoplay,loop,width=0.8\linewidth]{6}{frantic-}{0}{5}
    \end{frame}
    }
    \caption{Gif responses in conversation like the one shown above are embodied dialog that use visual imagery to convey reactions and emotions. This paper develops a system to select the appropriate gif response to messages. (PDF best viewed with Adobe Acrobat) }
    \label{fig:intro-example}
\end{figure}

Conversation analysis is central to NLP and multiple approaches have analyzed this dialog structure \cite{jurafsky1998lexical,pareti2018dialog,cohn2019large} and developed conversational agents to engage with people \citep[e.g.,][]{fang2018sounding,xu2020conversational,hong2020audrey}. Recent work has focused on generating open domain social chatbots that engage in sustained conversations in a natural way \cite{ram2018conversational}. Because many of these systems are designed to support voice-based dialog, they overlook non-textual forms of interaction used in social media conversations. In parallel, multimodal NLP systems have been developed for image data, often focusing on image-to-text tasks such as image captioning \cite{melas2018training,sharma2018conceptual} and visual question answering \cite{antol2015vqa,huang2019multi,khademi2020multimodal}. More recent work has focused on the reverse text-to-image dimension, such as generating an image from a description \cite{niu2020image,ramesh2021dalle}. Our work unites these two strands of research by integrating image-based communication into conversational agents.

Our paper offers three main contributions. 
First, we propose the new task of selecting gif responses in multimodal conversation analysis and introduce a new dataset of 1,562,701 real-world conversation turns with gif replies. 
Second, we introduce a new model \textsc{Pepe the King Prawn} that fuses image and text-based features to select a relevant gif response. In in-house experiments, we show that our model substantially outperforms strong baseline models at selecting the exact gif used in real data and, in a manual test of the quality of the best responses, achieves an nDCG of 0.8145 on the annotated test set.  
Third, in a real-world test, we deploy our model as a part of a large-scale randomized controlled trial and show that the gif replies produced by our model are more highly voted by the community. Data, code, and models are available at \href{https://github.com/xingyaoww/gif-reply}{https://github.com/xingyaoww/gif-reply}.

\section{GIF Communications}

Gifs have been widely adopted in communication as a natural form of embodied speech where the visual imagery conveys emotions or a reaction as a response \cite{bakhshi2016fast,tolins2016gifs}. These gifs commonly come from widely-known cultural products, such as movies or television shows, which provides common knowledge for how they could be interpreted \cite{eppink2014brief,miltner2017never}. However, a single gif may have multiple interpretations, depending on the context, cultural knowledge of its content, and the viewer \cite{jiang2017understanding}. As a result, a single gif can serve multiple functions in communication \cite{tolins2016gifs}.

Gifs have grown in their use through increasing affordances by platforms like Tumblr, Reddit, Imgur, and Twitter that allow gifs to be natively displayed like text in conversation threads \cite{jiang2018perfect}. Further, gif-based keyboards have been introduced that allow users to search for gifs that have been tagged with keywords or other metadata \cite{griggio2019customizations}. Yet, these technologies require that gif data be prepared with sufficient tags to be searchable or to have sufficient data to use collaborative filtering techniques for recommendations \citep[][p.9]{jiang2018perfect}. As a result, there is a clear gap in identifying appropriate response gifs directly from the text, which this work fills.

\section{Data}
\label{sec:data}

Despite the widespread use of gifs, no standard dataset exists for text and gif replies. Further, although platforms like Twitter support gif replies, these gifs are not canonicalized to identify which responses correspond to the same gif. Therefore, we construct a new dataset for this task by collecting responses, matching their images, and augmenting this data with metadata about the gif, where possible.
A visual description of the whole procedure can be found in Appendix \fref{fig:data-pipeline}.

\subsection{Gif Response Data}

Gifs have many uses \cite{miltner2017never} and so we use a two-step approach to collect data that focus specifically on those likely to be used in conversation.
First, gif responses are collected from Twitter by identifying all replies to English-language tweets containing \texttt{animated\_gif} as embedded media. Tweets were collected from a $\sim$10\% sample of Twitter from March 13th, 2019 to Jan 24th, 2020, totaling 42,096,566 tweets with a gif that we were able to retrieve.
Twitter does not canonicalize its gifs so two separate gif files may actually have the same imagery. Further, these files may not be identical due to small differences such as color variations or aspect ratios. To identify uses of the reference gifs, we use Average Hash from the \texttt{imagehash} library to create low-dimensional representations of each gif where hash distance corresponds to perceptual distance. Since gifs are animated and may contain varying scenes, we compute the hash for the first, middle, and final frames, concatenating these into a single hash.
Two gifs are considered the same if (i) they have identical hashes or (ii) their hamming distance is $<$ 10 and gifs with that hash have been used more than 500 times in Twitter. This latter condition was selected after manual evaluation of thresholds to trade-off between increasing the size of the training data and reducing potential noise caused by matching error. A visual example of this process can be found in Appendix \fref{fig:AverageHashExample}.

Not all gif responses in the Twitter data are conversational or appropriate for wider re-use. Therefore, we filter these responses to only those gifs whose imagery matches gifs hosted by the Giphy website, which is the backend for many gif-based keyboards. Giphy contains a wide collection of gifs that are curated to remove content inappropriate for general use (e.g., violent or sexual imagery). Gifs on the platform are categorized (e.g., ``reaction'' or ``celebrities'') and we identify 28 categories containing 972 keywords likely to contain gifs used in conversation. 
A total of 2,095,993 gifs linked to those keywords were ultimately retrieved and stored as image hashes. Additional details of categories and keywords are in Appendix \ref{appendix:GIFCategories}.%

\begin{figure} 
\centering
\includegraphics[width=0.43\textwidth]{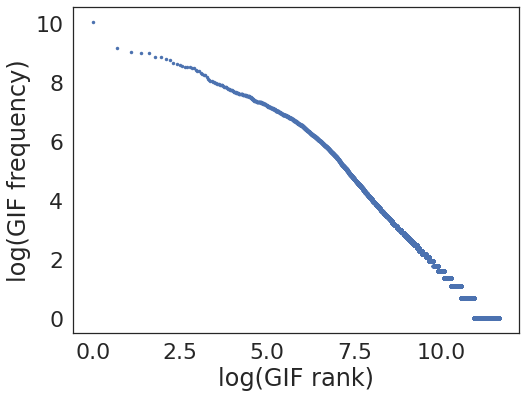}
\caption{The frequency distribution of gifs in our data roughly follows a log-normal distribution, with a few gifs used often, while a long tail of gifs are used relatively infrequently.}
\label{fig:zipf-law}
\end{figure}

After the matching image hashes to filter replies, we identify 115,586 unique gifs, referred to as \textit{reference gifs}, and 1,562,701 tweet replies using one of these gifs, which forms our official dataset. \fref{fig:zipf-law} shows these gifs' frequency in the data; much like words, a handful of gifs receive widespread use, while a long tail of gifs are rarely used.

\subsection{Gif Metadata}
\label{sec:GifMetadata}

We augment our gif data with  information about their content. Some gifs have text that transcribes what a person is saying in the gif's scene or is a meta-commentary on the content.  This text is extracted using paddleOCR  \cite{du2020paddleocr}. Since some gifs are long enough to contain multiple utterances, we run OCR on four frames sampled from each quartile of the gif's length. 
Roughly 50\% (58,020) of gifs contain at least one extracted word from the selected frames, with an mean of 5.5 extracted words per gif across the dataset.

Second, some gif repositories like Giphy allow users to tag gifs with information on their content or theme, e.g., ``face palm'' or ``movie.'' We collect tags for the 115K reference gifs used in Twitter, obtaining 39,651 unique tags. These user-generated tags were moderately noisy due to orthographic variations like spelling, capitalization, and spacing. Therefore, we merge tags by (i) lower-casing the text and (ii) performing a manual merge for similar word forms (e.g., ``excited'' and ``exciting''). To minimize noise, we retain only tags that have been used with at least five gifs and where those gifs have been used at least 1000 times in total; this process removes many low-frequency tags that are either overly-specific or idiosyncratic in their use.

Finally, we performed a manual inspection of all remaining tags to remove tags that are too general (e.g., ``emotion'') and retain only noun, adjective, and verb tags (words or multi-word expressions)  that describe specific emotions or actions. A total of 241 unique tags were retained (Appendix \ref{appendix:SelectedTags}).  6.0\% of gifs have at least one tag associated with them (mean 1.9 tags). However, these tagged gifs account for 38.7\% of the replies in our dataset, suggesting tags are only available for more-popular gifs.
Our dataset represents roughly an order of magnitude more data and more tags than the closest related dataset of \citet{chen2017gifgif} that contained 23K gifs with 17 manually-curated emotions.

\section{Gif Reply Models}
\label{sec:models}

We introduce a series of models for producing a gif response in conversation. Each model will select a gif from the 115K gifs in our dataset as a response to a text-based message. This task is related to but distinct from work on image-text matching \cite{Lee2018StackedCA}, which aims to find an image describing a piece of text, or text-to-image \citep[e.g.,][]{wen2015omg, Xu2018AttnGANFT}, which generates an image from a text description. Here, we aim to select gifs that reflect natural continuations or reactions to a message in a dialog, akin to how gifs are used in social media. For all models, additional details on the training procedures and hyperparameters are provided in Appendix \ref{app:model-training}.
The three models that follow use varying degrees of information about the gifs and text to select a response.

\subsection{Tag-based Predictions}

The first model uses tags as a shared representation for characterizing gifs and text. Analogous to how object tags are used as anchor points for image-text matching \cite{li2020oscar} and pivot languages are used in machine translation \cite{Cheng2017JointTF}, we use tags to bridge information between the text in a tweet and the visual content of a gif. Here, each gif becomes associated with a set of tags describing its conversational functions and for each text, we predict the set of tags for gifs responses to it---in essence, predicting what types of responses are most appropriate. We describe both of these processes next and how gifs are ultimately selected.

\begin{figure*}[t]
\begin{center} 
\includegraphics[width=\linewidth]{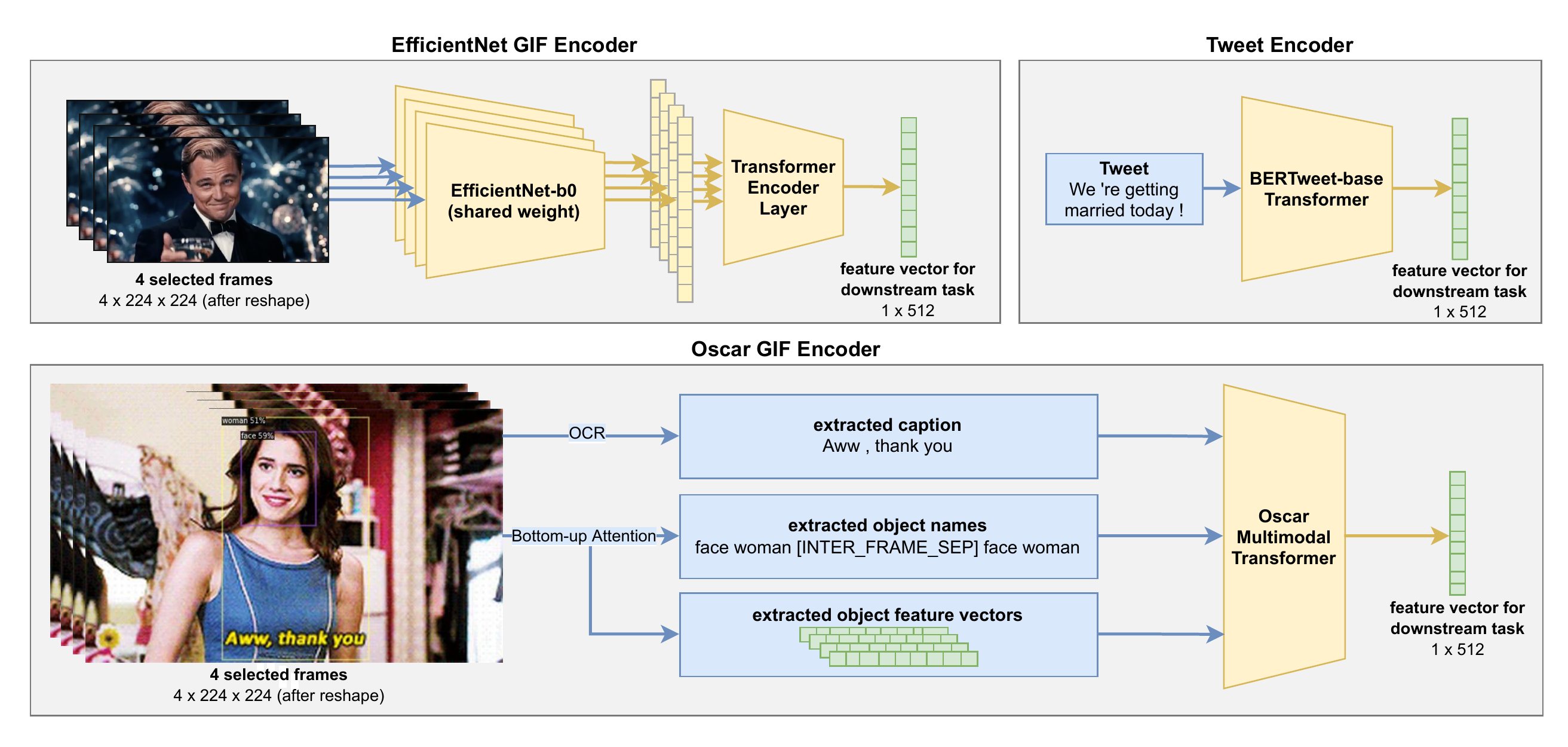}
\caption{The different encoder modules used to construct the models in \Sref{sec:models}.}
\label{fig:model-architecture}
\end{center}  
\end{figure*}

\myparagraph{Estimating Gif Tags}
Only 6.0\% of the gifs in our data have associated tags. Therefore we train a neural model to predict tags using known tags as training data. To capture any changes in emotion or imagery across the gif, we make separate predictions for four frames sampled across the gif (the same used in \Sref{sec:GifMetadata}).
Each frame is passed through an EfficientNet-based \cite{tan2019efficientnet} GIF encoder, shown in \fref{fig:model-architecture}, to extract a low-dimensional feature vector from each frame. These frame embeddings are fused using the attention mechanism from a transformer encoder layer. The output of the transformer feeds into a fully connected layer, which is trained as a multi-label classifier using binary cross-entropy to predict which tags should be present.

\myparagraph{Predicting Response Tags for Text}
For each message, we predict the $k$-hot distribution of tags for a gif response by training a BERTweet model \cite{nguyen2020bertweet}, which has been pre-trained on a large corpus of Twitter data (shown as  ``Tweet Encoder" in Figure \ref{fig:model-architecture}). The model with an additional fully connected layer is trained as a multi-label classifier using binary cross-entropy, using the tags for the gifs used in reply (if known).  

\myparagraph{Tag-based Gif Selection}
At inference time, given a message, we use the text-to-tag model to predict a \textit{k}-hot distribution over tags. Then, we select the gif whose estimated tag distribution is closest in Euclidean distance.

\subsection{CLIP variant}

The second model uses an end-to-end training approach based on the architecture of OpenAI CLIP  \cite{radford2021openaiclip}.
The architecture features two encoders, one for text and one for images. During training, the encoders are updated using contrastive loss that maximizes the cosine similarity of paired image-text representations and minimizes the cosine similarity of random pairs of images and texts. We replicate the CLIP architecture and training procedure, using BERTweet to encode text and EfficientNet \cite{tan2019efficientnet} to encode a composite image of four frames from the gif (compared with BERT and ResNet in their implementation).
While originally designed to select an image for a text description, our model is trained to select a gif reply for a text message---a more challenging task than the image retrieval task used in the original CLIP setup, as the message may not contain words describing elements of the gif. 
At inference time, given a tweet, we use the trained tweet encoder to extract its representation and compute its cosine similarity with each encoded representation for our gifs. The gif with the highest cosine similarity is returned as the best response.

\subsection{\textsc{Pepe the King Prawn}}

Our final model, \textsc{King Prawn}\footnote{\textsc{King Prawn} refers to ``selecKting INteresting Gifs for Personal RespAWNses.'' In this crazy muppet-name-land-grab world we live in, our only regret is that we couldn't get ``Pepino Rodrigo Serrano Gonzales'' to fit as a bacronym, which we leave to future work.} (referred to as ``\pepe''.) selects gif responses by using a richer set of multimodal features to create a gif representation. Rather than encode the gif solely from its image content, we use a multimodal encoder that captures (i) any text it might have, (ii) the types of objects present in the gif, and (iii) object regions as visual features. We encode these gif aspects using an \textsc{Oscar} transformer \cite{li2020oscar} to create a unified representation, shown in Figure \ref{fig:model-architecture} (bottom). Object names and regions of interest feature vectors are extracted using a pre-trained bottom-up attention model \cite{anderson2017up-down}. 

As input to the \textsc{Oscar} encoder, the captions to each of the gif's four frames are concatenated together with an ``[INTER\_FRAME\_SEP]" separator token. We filter object areas detected by the bottom-up attention model \cite{anderson2017up-down} and we keep all objects with probability $>$0.5. We then concatenate object names together with the same inter-frame separator between names of different frames. Together, the caption text, object names, and image-region features are fed into the \textsc{Oscar} transformer encoder to generate a GIF feature vector; the transformer is initialized with the default \textsc{Oscar} weights.
We use  BERTweet to encode text. 
The entire \pepe model is trained end-to-end using contrastive loss, similar to the CLIP model.

\section{Evaluation}

We initially evaluate the methods in two ways. First, we use traditional classification-based evaluation, testing whether the models can reproduce the observed gif replies. However, some messages could have multiple valid gif responses. Therefore, as a second test, we evaluate the model in a retrieval setting, measuring whether its most-probable responses are good quality for a message.  

\myparagraph{Experimental Setup}
Models are trained and tested on a dataset containing 1,562,701 Tweet-GIF pairs associated with 115,586 unique gifs, where 605,063 tweet-gif pairs are associated with at least one tag. Using the finalized 241 unique tags as classes for multi-label classification, we split the dataset by stratify on tags using the iterative train-test split method provided by \texttt{scikit-multilearn} library \cite{sechidis2011stratification,szymanski2017network} to create a 80:10:10 train, dev, and test split which is finalized to train the models described in \Sref{sec:models}.
Following BERTweet \cite{nguyen2020bertweet}, we preprocess tweets in our dataset using NLTK \texttt{TweetTokenizer} for tokenization, \texttt{emoji} package to translate emotion icons, and converted mentions and links to special ``@USER" and ``HTTPURL" tokens.

\myparagraph{Annotated Data}
To test whether each model's predictions are valid responses, we annotate the ten most-probable gif predictions for a subset of the tweets in our test data. Many tweets in our test set require substantial context to understand due to having few tokens, linking to URLs that provide extra knowledge, mentioning other users in directed communication. 
These factors suggest social context or general knowledge aids in the recipient's understanding of the gif’s intentions. While the model can still benefit from training on such examples, judging the appropriateness of response is difficult without access to the social context.
Therefore, to reduce interpretation ambiguity, we annotate only tweets without URLs or user mentions and having at least 10 tokens. This process selects tweets with sufficient content to judge appropriateness independent of the larger social context. 

Two annotators (the authors) were shown a list of potential gif responses for a tweet and asked to judge whether this is an appropriate gif response (a binary rating). Gifs were selected from the ten most-probable replies for each system and collectively shown in random order to prevent knowing which system generated each reply. A total of 2,500 gif-tweet pairings were annotated. Annotators attained a Krippendorf's $\alpha$ of 0.462; while moderate agreement, this value is expected given known differences in how people interpret and value gif responses based on their familiarity with its content, message interpretation, and life-experience \cite{jiang2018perfect}.
We follow the evaluation setup from other retrieval-based dialog systems \citep[e.g.][]{yu2021few,kumar2020making} and use normalized Discounted Cumulative Gain (nDCG), which measures whether more appropriate gif responses are ranked higher. A gif's appropriateness score is the sum of annotators' ratings. %

\begin{table}[t]
\centering
\resizebox{0.5\textwidth}{!}{
\begin{tabular}{lrrr}
    \textbf{Model} &      \textbf{Top-1} &      \textbf{Top-5} &     \textbf{Top-10} \\
    \hline
    Tag-based              &  0.000000 &  0.000092 &  0.000119 \\
    Random                 &  0.000020 &  0.000059 &  0.000158 \\
    CLIP variant           &  0.000488 &  0.001669 &  0.002783 \\
    Distribution sampling  &  0.000996 &  0.005098 &  0.009780 \\
    \pepe                  &  \textbf{0.005375} &  \textbf{0.018723} &  \textbf{0.030918} \\

    \end{tabular}
}
\caption{Models' precision-at-$k$  on selecting the \textit{exact} gif used as a response for a tweet in the test set; this performance is an underestimate of each model, as many model-predicted gifs may be appropriate. }
\label{tab:test-data}
\end{table}

\begin{table}[t]
\centering
\begin{tabular}{lc}
\textbf{Model} & \textbf{nDCG} \\
\hline
Random  & 0.3273 \\
Tag-based  & 0.4526 \\
Distribution sampling  & 0.4969 \\
CLIP variant  & 0.5934 \\
\pepe  &\textbf{0.8145} \\

\end{tabular}
\caption{Models' nDCG scores at proposing appropriate gif replies, measured from annotations on the top 10 most probable gif replies of each model.}
\label{tab:annotated-data}
\end{table}

\begin{table}[t]
\centering
\begin{tabular}{lc}
\textbf{Model} & \textbf{nDCG} \\
\hline
\pepe & \textbf{0.8145} \\
\pepe without object names  & 0.7665 \\
\pepe without caption  & 0.7559 \\
\pepe without object features & 0.7533 \\
\end{tabular}
\caption{Results for ablated versions of \pepe where specific input is removed (cf.~\tref{tab:annotated-data}) show that all input forms contribute to the ability to select replies.}
\label{tab:ablation-annotated-data}
\end{table}

\begin{table*}[t]
\resizebox{1\textwidth}{!}{

\begin{tabular}{m{0.22\textwidth}p{0.15\textwidth}p{0.15\textwidth}p{0.15\textwidth}p{0.15\textwidth}p{0.15\textwidth}}
\textbf{Parent Tweets} & \textbf{Tag-based} & \textbf{CLIP variant} &\textbf{\pepe} & \textbf{Dist. Samp.} & \textbf{Random}  \\ 
\hline
That wonderful feeling you get when you arrive to a business dinner that you're supposedly paying for...and realize you've forgotten your credit card  &  \vgif{DbDBUaPPBux7a} & \vgif{JUMLTR3dHEGpW} & 
\vgif{91X2MlgF7dHsA} & \vgif{q4sdF9tchap6E} & \myvgif{PdkAipBoyAkc8} \\ 

I'm convinced some of y'all don't get laid & \vgif{3ohze1C8xLTsqxEye4} & \vgif{loitbnzQ1JQ8Iizx8w} & \vgif{PLZBEW6h1cEhlgsImy} & \vgif{G4ZNYMQVMH6us} & \vgif{25EAreLnZIFWasblJe} \\
\end{tabular}
}
\vspace{-3mm}
\caption{Model-selected replies to messages (paraphrased for privacy). Click an image to view the gif on Giphy.}
\label{tab:model-reply-example}
\end{table*}

\myparagraph{Results}
The \pepe model was able to identify relevant and good-quality gif responses, as shown by its performances on the test data (\tref{tab:test-data}) and annotated data (\tref{tab:annotated-data}). Performance on the test set is expected to be low, given the challenge of identifying the exact gif used for a tweet when multiple possible gifs are likely to be equally valid. However, the \pepe model is still able to identify the exact gif (out of 115K) in its top 10 predictions for  3\% of the data, substantially outperforming all other models.

Performance on the annotated data (\tref{tab:annotated-data}) provides a more realistic assessment of whether models can generate high-quality replies, as it measures whether the models' replies themselves were good. The \pepe model attains substantially higher performance (p$<$0.01) than other models. While the CLIP variant model performs well, the content-agnostic Distribution sampling baseline performs nearly as well. This baseline's high performance speaks to the multiple interpretations of gifs and the ease at which readers can make connections between a gif and message. Indeed, even the random-gif model has a non-zero nDCG, highlighting the ability for an arbitrary gif to still be considered appropriate. We speculate that popular gifs may be popular because of this ease of multiple interpretations.
\tref{tab:model-reply-example} shows the top predictions for models and baselines for two example messages, illustrating the variety of relevant gifs; the \pepe and random baseline replies for the second message exemplify the type of gifs that can be widely applied to many messages, often to humorous effects.

\myparagraph{Ablation study}
\pepe fuses multiple types of input, which may uniquely contribute to model's ability to select gif replies. To  understand how these inputs each contribute, we performed an ablation study on the annotated test set by removing one input from Oscar GIF Encoder shown in \fref{fig:model-architecture}  (i.e., a gif's caption, object names, or objects' visual features) and evaluating the model's resulting gifs on the same test instances.

The ablated model performances, shown in \tref{tab:ablation-annotated-data}, reveal that each input is useful for selecting gifs.\footnote{The performance decrease for removing object names is statistically significant (p$<$0.01, bootstrapped). The decreases for removing captions and objects' visual features are significant from the name-removal model (p$<$0.01) but the two models are statistically equivalent (p$>$0.19).}
Object features capture visual information about what specifically is present in the gif (beyond the discrete names of what is present, e.g., ``person'' or ``building'') and show that multimodality is important for high performance---predicting replies just from a gif's caption and categorized content are insufficient. Similarly, the caption of a gif (if present) is important, as the text can help make explicit the intended interpretation of a gif.

\section{Field Experiment}

To test the generalizability of our models and quality of their responses, we conduct a large-scale randomized controlled trial (RCT) that has the models respond to real users and measure their perception of reply quality.\footnote{This experiment was ruled as Not Regulated by the University of Michigan IRB (HUM00197631). However, IRB approval is not sufficient to prevent harm \cite{bernstein2021esr} and significant precautions were taken to minimize potential risk (See \Sref{sec:ethics}) . }

\subsection{Experimental Setup}

Gifs were posted to the Imgur platform, which is a highly active social media community that supports both image and text-based interactions. On Imgur, users may create posts, which contain one or more images with optional commentary, or comment on posts or replies. Similar to pre-2018 Twitter, comments are limited to 140 characters. Imgur conversations are threaded and frequently contain both image and text comments. Like Reddit, users may upvote and downvote content, providing a score of how well it was received by the community; we use this score in our experiments to evaluate quality.

Our experiment focuses on generating Gif-based replies to top-level text comments (comments made directly to the post). This setup mirrors the conversational data our models were trained on.
Imgur supports several ways of filtering its stream of posts. %
To ensure that our replies have sufficient visibility, we select posts that have already receive 10 comments and appear in the ``most viral'' sorting.
From these posts, we reply to the top-rated text comment.  The RCT runs from 8 AM to 8 PM (local time), making at most 10 replies per hour.

Not all topics or comments are suitable for automated responses and great care was taken to prevent potential harm to the community. Through multiple rounds of testing which replies would be responded to, we curated a list of keywords that could lead to potential controversial replies, such as terms about religion or race (full list in Appendix \ref{appendix:ImgurFilterList}). Any comment containing a token or lemma matching a word on this list is excluded and not replied to. As a further safeguard, experimenters monitored all replies to remove any that were deemed inappropriate. See the Ethics Section (\Sref{sec:ethics}) for a longer discussion of safeguards.

The field experiment consists of five arms, corresponding to the three trained models and the two baseline models. During each trial, one model is selected and generates a response; the trained model replies with the most probable gif.\footnote{Due to a bug, early experimental trials for the CLIP and \pepe models used the tenth most-probable gif; however, using the ratings in the annotated data, a $t$-test of the difference in quality for most- and tenth-most probable gifs showed no statistically-significant difference in quality for both models (p>0.1). Therefore, we include this data in our results.}

Not all models are equally likely to perform well and so to make the most use of our trial budget, we use Thompson sampling \cite{russo2018tutorial} to randomly select which arm of the trial to use. Thompson sampling builds a probability model for the estimated reward of each arm (here, the score a reply receives) and samples from the model such that higher-rewarding arms are sampled more frequently. As a result, this method can provide tighter estimates for the reward of the most useful arms. Scores in Imgur have a skewed distribution, with few comments receiving very high scores and most receiving near the default score (1). Therefore, we use Poisson Thompson sampling. Some comments may be downvoted to receive scores below zero, so for simplicity, we truncate these scores to 0.

We initialize the reward estimates for our experiment by selecting one of the five models in a round-robin manner to reply to an Imgur comment for 3 days. These initial scores act as priors for Thompson sampling to update Poisson distributions for each model. In the trial, we choose a model by sampling from the up distributions using all previous days' scores as the prior.
The experiment ran from April 15th, 2021 to August 30th, 2021, and models generated a total of 8,369 replies.

To evaluate the results of the RCT, we construct a Negative Binomial regression on the dependent variable of the score received for a model's reply, truncating negative scores to zero. The Negative binomial was chosen instead of Poisson due to over-dispersion in the score variable.  The models are treated as a categorical variable, using the random model as a reference. Since the score will depend, in part, on the attention received by the parent post and comment (higher-rated comments are displayed first), we include linear effects for the post and parent comment. Finally, we include five text-related variables to control for the content of the parent comment: the topic distribution (Appendix~\tref{table:topic-model-keywords}) from a 10-topic model (dropping one topic due to collinearity), the sentiment and subjectivity of the message estimated using \texttt{TextBlob} library, the length of the comment, and  whether the comment contained a question. %

\subsection{Results}

\begin{figure}[t]
\centering
\includegraphics[width=0.44\textwidth]{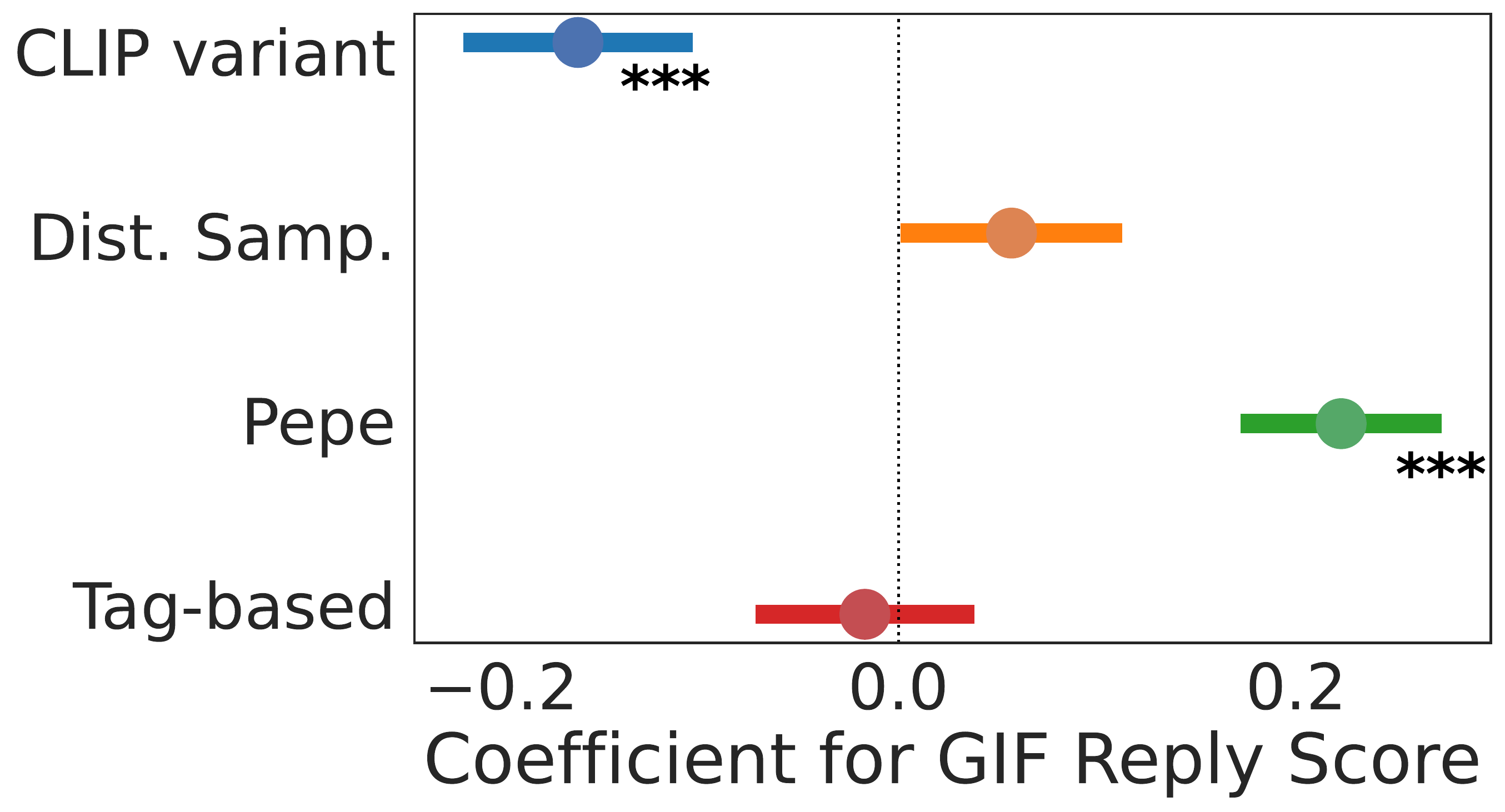}
\caption{Negative Binomial regression coefficients for each model on predicting a gif reply's score, using the random-gif model as the reference category; bars show standard error and *** denotes significance at 0.01.}
\label{fig:RegressionPlot}
\end{figure}

The field experiment demonstrates that the \pepe model is able to generate significantly higher-scoring responses. \fref{fig:RegressionPlot} shows the Negative Binomial regression coefficients for the three models and empirical distribution baseline, with the random gif model as a reference; full regression results are shown in Appendix \tref{table:ScoreRegression}. The \pepe model substantially outperforms all other models (p$<$0.01) in this real-world setting. Surprisingly, despite performing second-best in our annotated evaluations, the CLIP model performs worst, with its replies receiving fewer upvotes than the two baselines that randomly select gifs. We investigate potential explanations for these performances next.

The Random and Distributional-sampling baseline models perform surprisingly well relative to models that take the text and gif content into account, with only the \pepe model outperforming them. The performance of the random baselines matches prior work showing people are still able to draw some connection between their interpretation and the reply \citep[][p.29]{madden2018phenomenological}. Further, we observed that, when the model's reply truly seemed random, some users replied say they upvoted solely because they enjoyed the gif. %

\begin{figure}[t]
\centering

\subfloat[Tag-based]{
  \includegraphics[width=0.9\columnwidth]{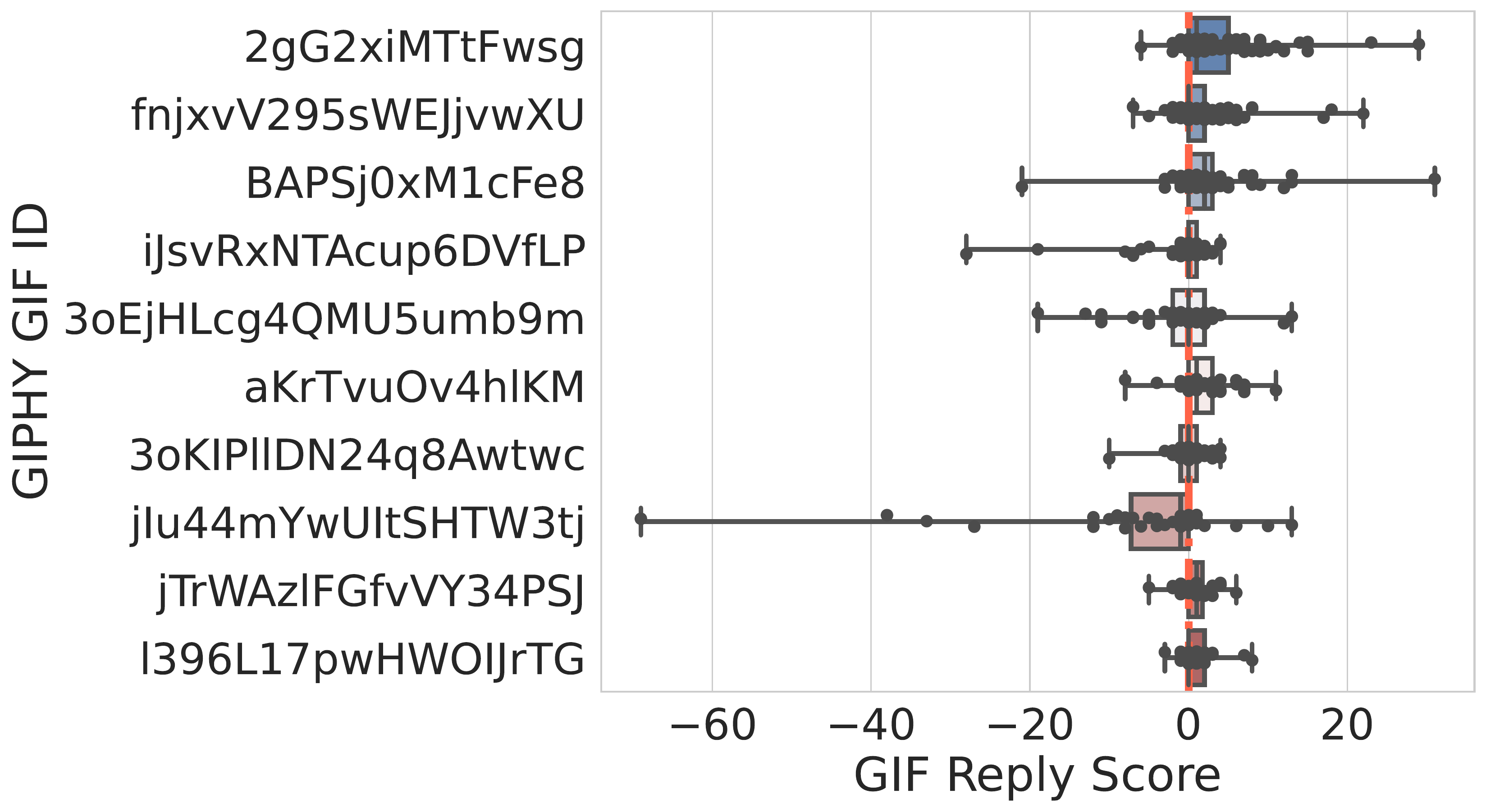}
}

\subfloat[CLIP variant]{
  \includegraphics[width=0.9\columnwidth]{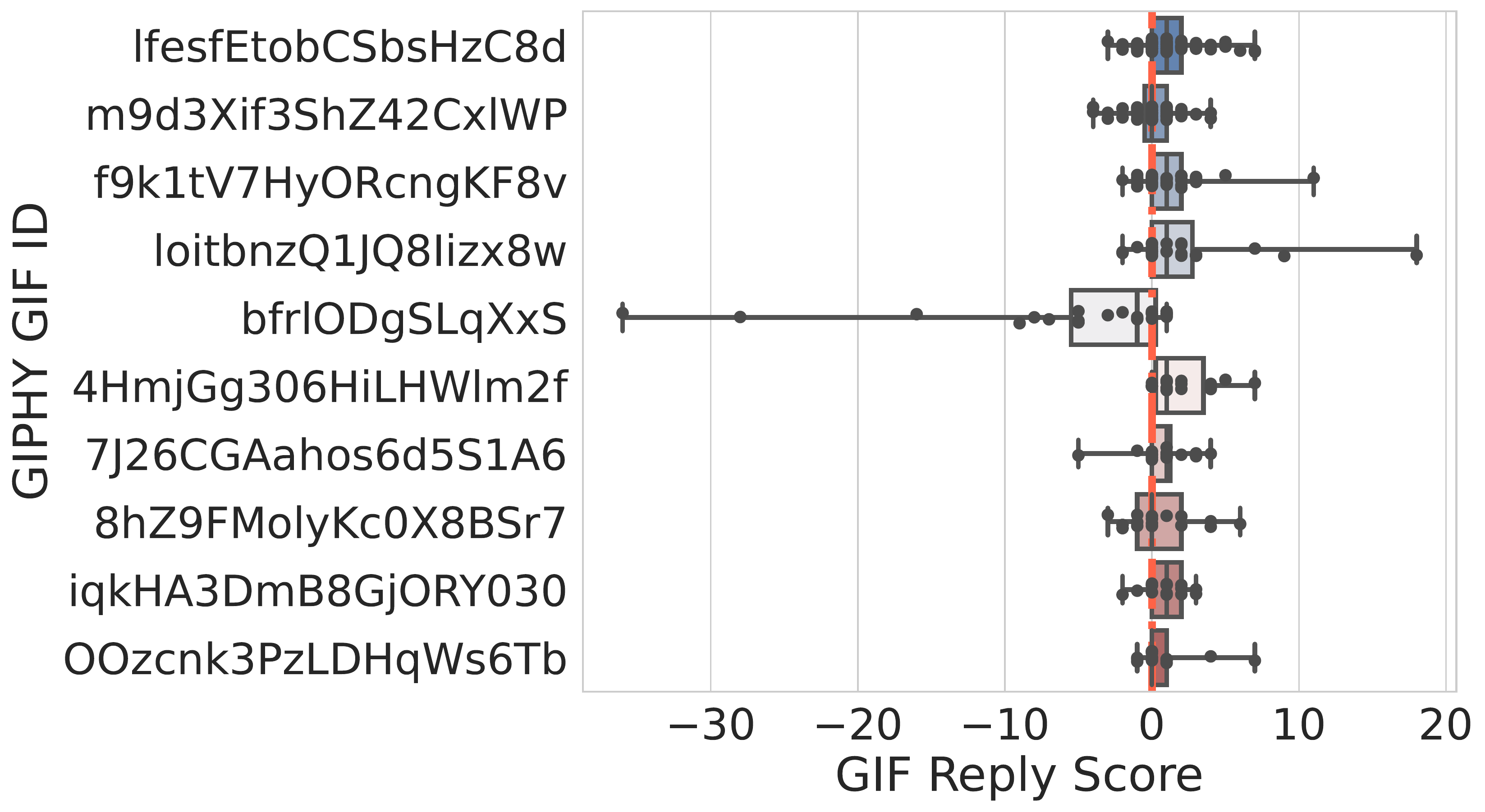}
}

\subfloat[\pepe]{
  \includegraphics[width=0.9\columnwidth]{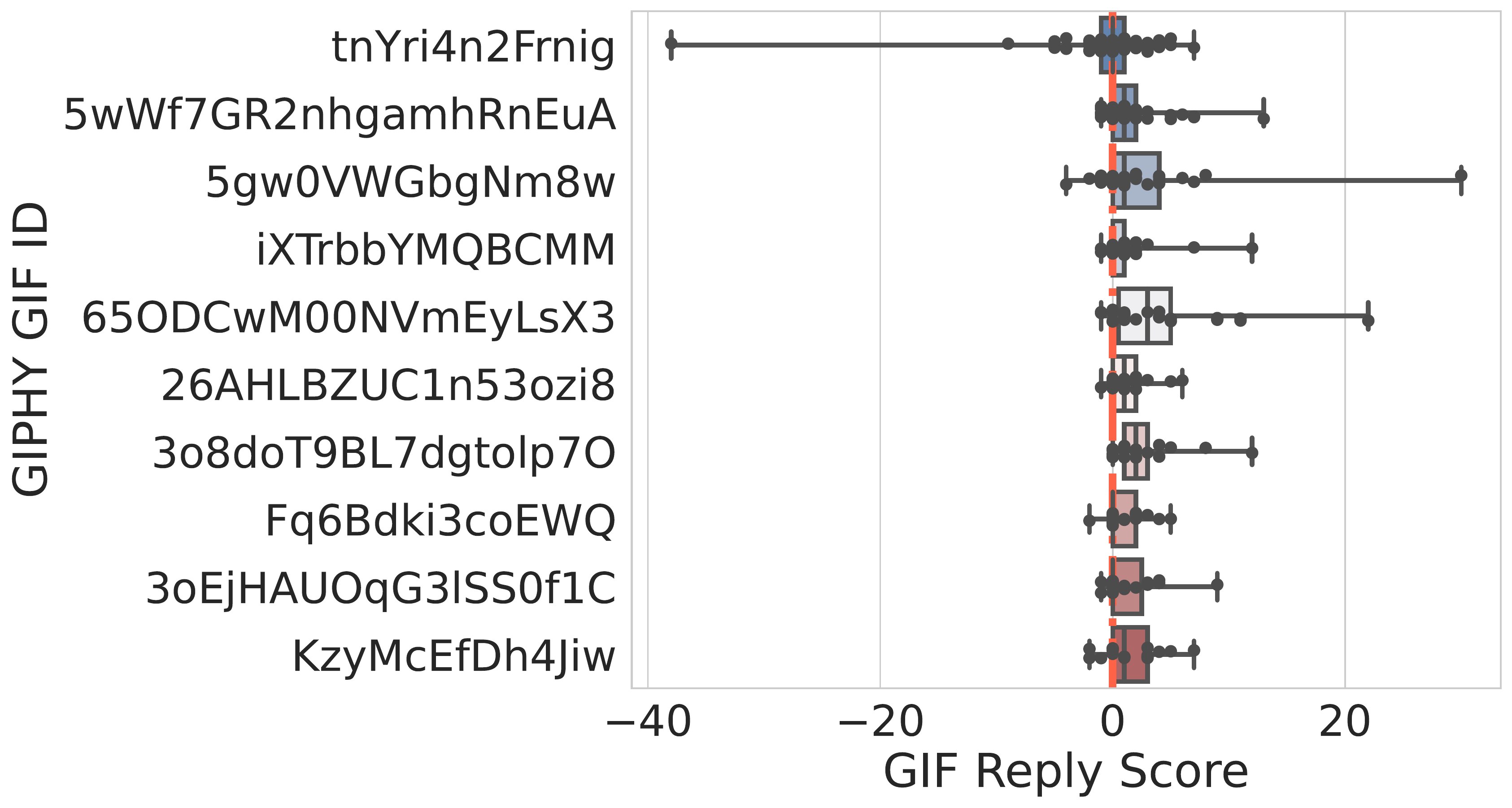}
}

\caption{Score distributions for most-frequently used gifs show few are universally skewed positive. Boxes show quartile ranges; gifs are in Appendix \tref{tab:top10-gif-example}. }
\label{fig:top10-gif-score}
\end{figure}

As a follow-up experiment, we tested whether models could be getting higher (or lower) scores by repeatedly picking the same gifs that are skewed towards a positive or negative reaction. Figure \ref{fig:top10-gif-score} shows the score distribution for the top ten most frequently used gifs (visual examples in Appendix~\tref{tab:top10-gif-example}) for each of the three trained models and reveals surprisingly divergent behavior for how the community reacts. 
Each model had a different set of most-used gifs, indicating the models did not converge to a universal set of common replies. Indeed, a gif's frequency-of-use and mean reply score were uncorrelated in all three models ($r\approx$-0.01, p$>$0.73 for all models).  The most-used gifs for each model had average scores that were positive, but the distributions for each gif show that some uses were occasionally downvoted. This high variance in scores indicates that a gif's intrinsic qualities are not solely responsible for the received score and, instead, appropriate use in context is plays a significant part in community reception.

\begin{figure}[t]
\centering
\includegraphics[width=0.48\textwidth]{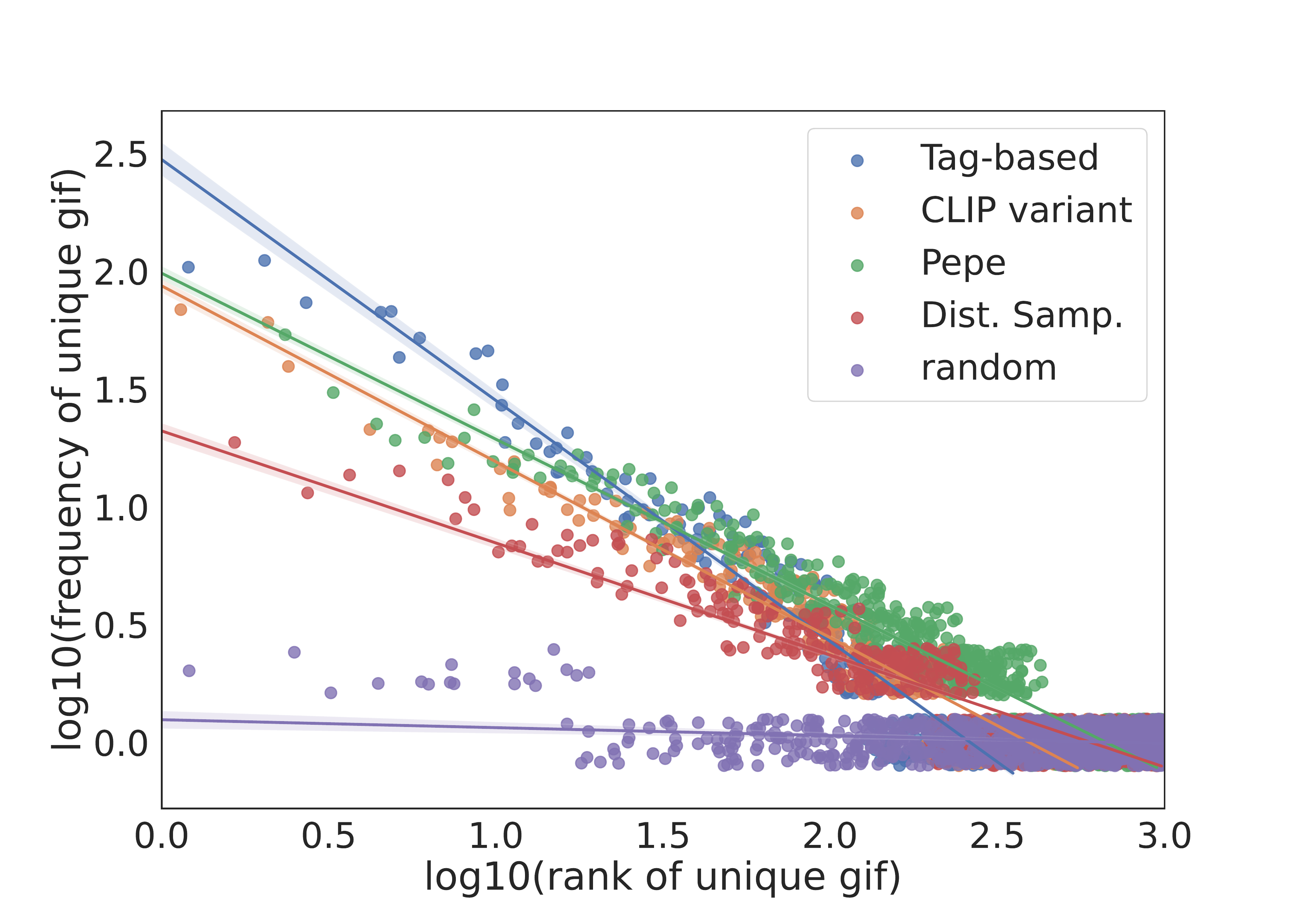}
\caption{Gif use frequency by each model, shown as frequency-vs-rank log-scaled with first-order line fit (jitter added for separation).}
\label{fig:model-zipf-plot}
\end{figure}

We examined whether models relied on the same set of gifs. \fref{fig:model-zipf-plot} shows the distribution of gif uses by each model, indicating that the tag-based model relied frequently on a small set of gifs. However, the \pepe and CLIP variant models were substantially more varied, indicating they draw from the long-tail of possible gifs.

Do any of our models spark more subsequent conversation? We fit a separate Negative Binomial regression on the total number of comments made to our reply, using the same IVs as the score regression and include the reply's score itself as another IV. This model (Appendix \tref{table:num-replies-regression}) shows that both the distributional-sampling baseline and \pepe models produced replies that led to \textit{fewer} subsequent comments (p$<$0.01)---despite the \pepe model having the most-upvoted replies. However, the score of the gif reply was positively associated (p$<$0.01) indicating that more appropriate replies do receive more subsequent conversation. We speculate that the random models may have led to more conversation due to users replying to express confusion about why the particular gif was used. This result points to a need to understand what text and visual factors in gifs influence the volume of subsequent dialog and an opportunity to optimize gif models for both quality and number of conversation turns.

\section{Related Work}

This work draws upon two strands of research from dialog systems and multimodal NLP. 
Conversational dialog systems have traditionally been built upon large-scale dialog corpora from social media platforms \cite{bessho2012dialog} such as Twitter. Our approaches are fundamentally information retrieval based systems that mirror the approach by text-based conversational systems that retrieve existing messages from a large social media corpus as potential replies and rank these to select a response. Our work mirrors models that use neural networks for ranking \citep[e.g.,]{yan2016learning,inaba2016neural,penha2021calibration}; however, we note that many recent knowledge-grounded and open domain models use encoder-decoder methods to improve versatility and applicability \citep[e.g.,][]{ghazvininejad2018knowledge,gao2019neural,zhou2020design}. Generative approaches are likely inappropriate for gif-based conversation as gifs are more akin to mimetic artifacts that build on cultural knowledge \cite{eppink2014brief}, making synthesizing a new gif from scratch likely less effective. 

All three models used here rely on joint embedding spaces for gif and text. Multiple works in NLP have been proposed to align these representations \cite{kiros2014unifying,wang2016learning}, often for particular applications such as visual question answering \cite{antol2015vqa}.  Recent work has focused on embeddings these media with a single encoder that takes both text and images as input \citep[e.g.,][]{wang2019camp,chen2019uniter}, in contrast to our model that uses separate image and text encoders (\fref{fig:model-architecture}); these multimodal encoders are prohibitively computationally expensive to use in our setting during inference time, as the model would need to be run on each gif (and message) to rank replies, compared with our model that only needs to encode text. However, performance and efficiency improvements in aligning image and text representations would likely benefit our task.

\section{Conclusion}

People like using gifs in online conversations---gifs are a fun and playful way to communicate. However, modern NLP conversational agents operate only by text. Here, we introduce a new dataset of 1.56M conversation turns using gifs, including captions and metadata, and develop a new conversational model \textsc{Pepe the King Prawn} that selects appropriate gif responses for messages through comparing encoded gif and text representations.  In two evaluations, we show that \pepe is able to generate highly-relevant gif responses  and in a large-scale RCT, we show that the gif replies from the \pepe model received significantly higher scores from the general public. Our work demonstrates the opportunity for using NLP methods to successfully engage in multimodal conversations.

\section{Ethics}
\label{sec:ethics}
The interactive nature of the RCT necessitated a close consideration of ethical issues \cite{thieltges2016devil}. Prior to beginning the RCT, the study team obtained IRB approval to interact with users. While necessary in the legal sense, IRB approval is not sufficient to justify the ethical grounds of the study. The primary risks of the study are if the automated models respond with an inappropriate gif or respond to a message that is not suitable for automated response (e.g., discussing the death of a loved one or making an offensive statement). These risks were mitigated in multiple ways throughout the dataset construction and field experiment.

First, the selection criteria for which comments we reply to was designed to only reply to content that was already deemed appropriate by the community. By selecting only posts that had received sufficient upvotes to be called ``viral'' and were already receiving comments, we mitigate the risk of engaging in topics or conversations that are inappropriate according to the norms of the Imgur community, as these posts would be removed by moderators or would have received sufficient downvotes to stay in obscurity. 

Second, by focusing on the top-voted comment to these posts, we again reply to content that has already been deemed high-quality by the comment. This comment-level criteria substantially lowers the risk of our models commenting on inappropriate comments (e.g., a comment insulting another user), as these comments are readily downvoted by the community prior to our intervention. 

Third, we employed extensive filtering to avoid replying to any comment containing a potentially sensitive topic, e.g., a discussion of race or trauma (keywords are listed in Appendix~\ref{appendix:ImgurFilterList}). The initial set of keywords was developed through examining potentially sensitive topics and then iteratively added to by simulating which messages our RCT would reply to and examining whether it would be appropriate. During the field RCT, experimenters continuously monitored the comments to ensure no harm was being done. Ultimately, only three comments were removed during the initial two days, which was due to a bug in the lemmatization and these comments should have been filtered out by our earlier criteria; these comments were removed quickly and we did not observe any notable response from the community.

Fourth, one risk is replying with an inappropriate gif, which is mitigated by the use of Giphy to seed our initial gifs. As this platform is curated and does not host objectively offensive gifs (e.g., overly-violent content), our initial gif set is relatively free of objectionable gifs. %
Because our model learns directly from gifs' frequency of use, unless objectively offensive gifs are widely used, they are unlikely to be deployed from our RCT; we speculate that few objectively offensive gifs are widely used  and, in practice, we have not identified any during the study period or when examining hundreds of random gifs in our data (or used in the RCT).

Finally, one risk is that by learning gif responses from observed data, our models may reinforce cultural stereotypes that are encoded in the gifs themselves \cite{erinn2019digital}, e.g., the association of African American individuals with strong emotions. While our gif data is relatively clean of overtly offensive gifs, we acknowledge that our model likely does inadvertently perpetuate some of these latent biases in the data. However, the success of our model suggests a future mitigation strategy for platforms suggesting gifs: as biases become known, our approach can be used to suggest less-biased gifs as potential responses to mitigate future harm.

\section*{Acknowledgments}

We thank the reviewers, area chairs, and senior area chairs for their thoughtful comments and feedback. We also thank the Blablablab for helpful feedback and letting us deploy \pepe to the group's Slack and putting up with the ridiculous gif replies and Imgur for being a wonderful community. This material is
based upon work supported by the National Science Foundation under Grant No.~2007251.

\bibliography{anthology,custom,references}
\bibliographystyle{acl_natbib}

\clearpage
\appendix

\begin{figure*} 
\begin{center} 
\includegraphics[width=\textwidth]{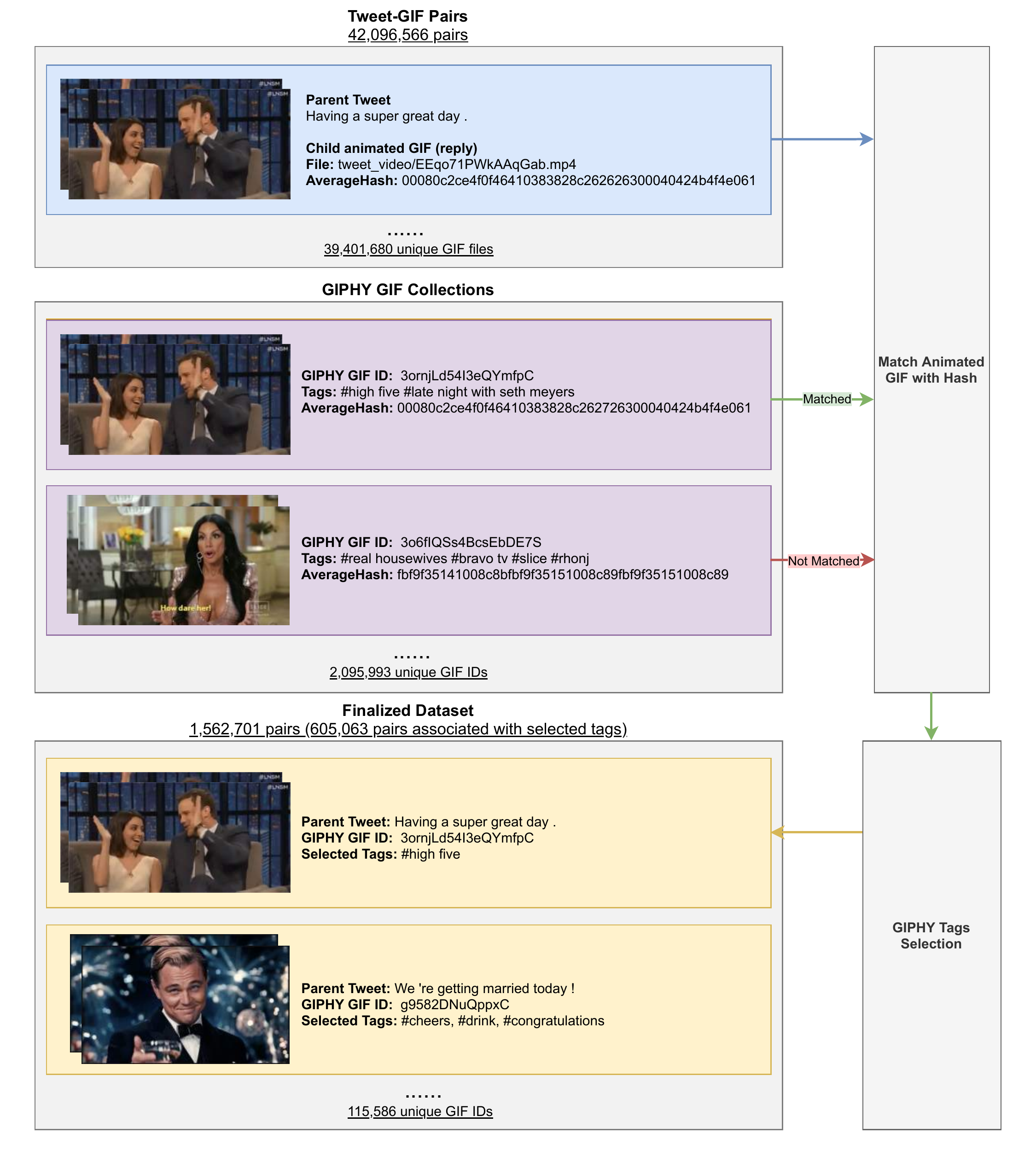}
\caption{A diagram of the pipeline used to collect, canonicalize, and filter gif-reply data from Twitter.}
\label{fig:data-pipeline}
\end{center}
\end{figure*}

\begin{table}[tb]
    \begin{tabular}{ll}
    \toprule \hline
    \textbf{Category} & \textbf{Subcategory} \\ \hline
    \midrule
        Cartoons \& Comics & aqua teen hunger force \\ 
        Celebrities &           richard pryor \\
        Reactions &             angry \\
        Emotions &              happy \\
        Anime &                 bleach \\
        Art \& Design &         psychedelic \\
        Nature &                sunrise \\
        Transportation &        bicycle \\  \hline
    \bottomrule
    \end{tabular}
    \caption{Examples of GIF categories on GIPHY}
    \label{table:GifCategoryExample}
\end{table}

\begin{figure*} 
\centering
\includegraphics[width=\textwidth]{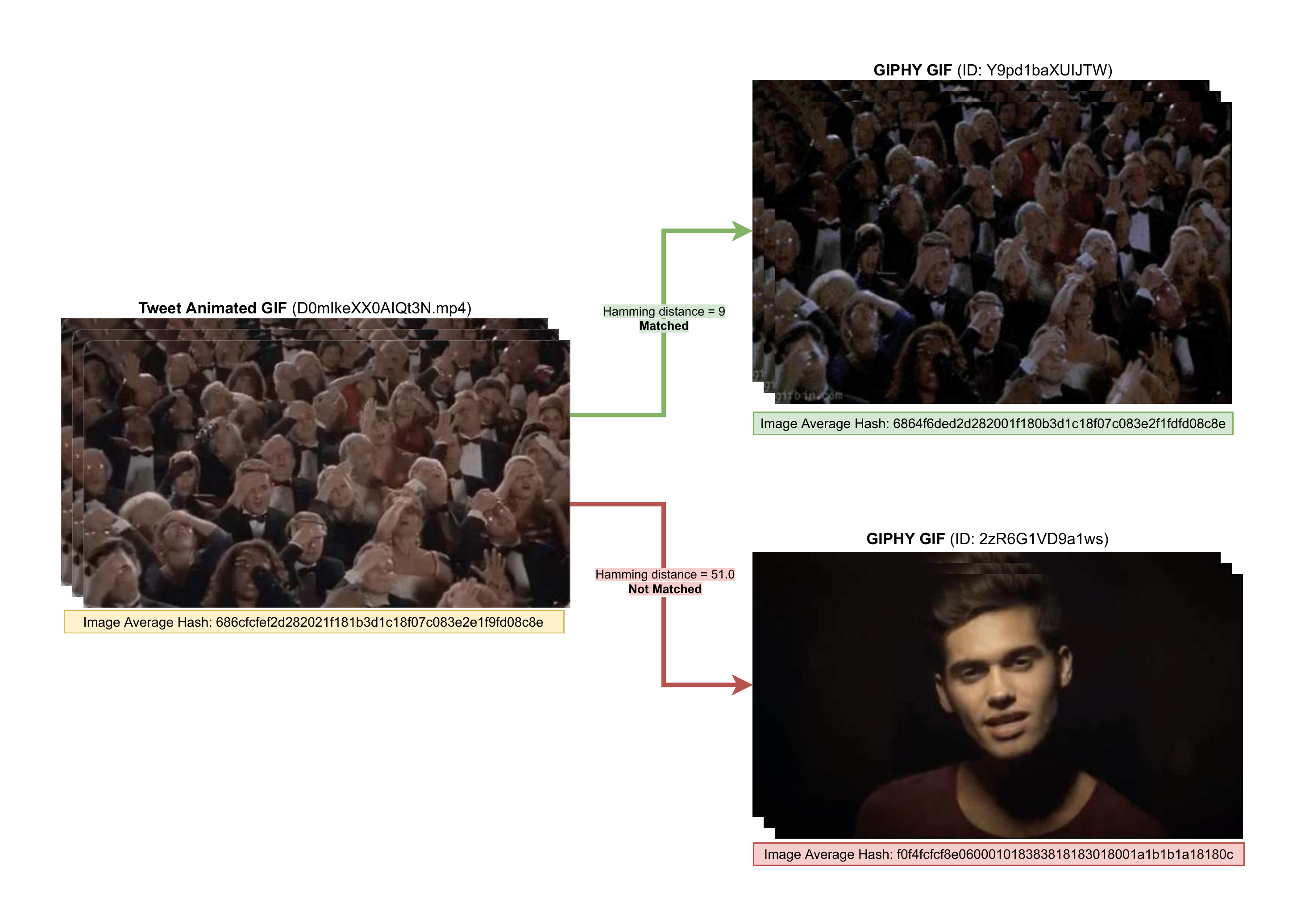}
\caption{Matching Animated GIFs from Twitter with GIPHY gifs using Image Average Hash}
\label{fig:AverageHashExample}
\end{figure*}

\section{Additional Details on Model Training}
\label{app:model-training}

Following, we provide additional details on how each of the three models was trained.

\subsection{Tag-based Model}

\myparagraph{EfficientNet-based Tag Classifier}
Gifs are reshaped to 224 by 224 pixel while keeping the aspect ratio by padding and normalized to a mean of 0.5 and standard deviation of 0.5 for each channel before feeding into the EfficientNet-based model.
We selected unique GIFs from the finalized dataset that has at least one associated tag and using the iterative train test split on k-hot tag representation to select 5\% of those GIFs for validation.
The EfficientNet tag classifier was trained for 100 epochs on a batch size of 32, using AdamW optimizer with learning rate 1e-5 and weight decay 1e-3. The best validation performance was achieved at the 40th epoch with macro-f1 of 0.30 in predicting 241 multi-label classes.
Early experiment shows that transformer encoder layer (macro-f1 of 0.30) out performs linear layer (macro-f1 of 0.19) in fusing multi-frame gif features on the development set, therefore transformer encoder layer is used to fuse features of different frames in our implementation.

\myparagraph{Tweet-to-tag classifier}
Using the finalized dataset mentioned in \Sref{sec:data}, we use tweet as input, and the k-hot tag representation of that tweet instance as ground truth label to train the multi-label classifier along with the tweet encoder for 241 classes. Additionally, we filter out tweets from the finalized dataset that do not have corresponding twitter tags before training. The model with the best validation performance is selected to perform subsequent evaluation and field experiments.
The tweet encoder was trained for 100 epochs with a batch size of 32. The learning rate was set to 1e-5 with 1e-3 weight decay using AdamW optimizer. The best validation macro-f1 was 0.07 achieved at the 70th epoch.

\subsection{CLIP variant}
The evaluation performance for model selection is measured by nDCG. For every tweet-gif pair in the validation set, we measure the top 30 predicted GIFs from the model using the tweet as input. The relevance of an occurring ground truth gif in the top 30 predictions given a tweet is set 1 for the nDCG calculation.

CLIP variant is trained on the same finalized dataset using contrastive loss. It was trained for 16 epochs with a batch size of 16 using AdamW optimizer of learning rate 1e-5 and weight decay 1e-3. Best validation performance is achieved at epoch 6 with an nDCG value of 0.015.

We replace the Transformer encoder layer with a linear Layer on Efficient GIF Encoder from \fref{fig:model-architecture}, and use this as our GIF Encoder for the CLIP variant. Image inputs to the GIF encoder are normalized following the official CLIP implementation.

\subsection{\pepe}

The \pepe model follows most configurations from the CLIP variant model, but replace the EffcientNet GIF encoder with an Oscar GIF encoder based on Oscar pre-trained multi-modal transformer \cite{li2020oscar}. 

Extra metadata are extracted from GIFs in the finalized dataset for further training. Captions within the GIF are extracted using PaddleOCR \cite{du2020paddleocr}, and only extracted text with probability greater than 0.9 are kept as caption metadata. 

Object tags and their corresponding features are extracted with bottom-up attention \cite{anderson2017up-down} using \texttt{py-bottom-up-attention} package. Object instances are filtered to only keep instances that have a score higher than 0.5, then object tags and their corresponding features are extracted from these instances. Final object features of dimension 2054 are obtained by concatenating feature output with dimension 2048 from Faster-RCNN with scaled box position coordinates of the object following \cite{li2020oscar}.

The \pepe model is trained on the finalized dataset with extracted caption and object metadata. It was trained for 16 epochs with a batch size of 8 using AdamW optimizer of learning rate 1e-6 and weight decay 1e-3. Preprocessing for GIFs is the same as the Tag-based model. Max sequence length is set to 256 tokens for the Oscar transformer. Best evaluation performance is achieved at epoch 12 with an nDCG score of 0.007.

\begin{table*}[!htbp] \centering 
\begin{tabular}{@{\extracolsep{3pt}}lc} 
\\[-1.8ex]\hline 
\hline \\[-1.8ex] 
 & \multicolumn{1}{c}{\textit{Dependent variable:}} \\ 
\cline{2-2} 
\\[-1.8ex] & Gif reply score \\ 
\hline \\[-1.8ex] 
 post score & $-$0.0002$^{***}$ (0.00003) \\ 
  comment score & 0.001$^{***}$ (0.0001) \\ 
  CLIP variant model & $-$0.161$^{***}$ (0.058) \\ 
  Distribution-sampling model & 0.057 (0.056) \\ 
  \pepe model & 0.223$^{***}$ (0.051) \\ 
  Tag-based model & $-$0.017 (0.055) \\ 
  number of days after reply & 0.003$^{***}$ (0.0005) \\ 
  comment text polarity & $-$0.039 (0.058) \\ 
  comment text subjectivity & $-$0.033 (0.052) \\ 
  topic 0 (Politics related) & 0.078 (0.155) \\ 
  topic 1 (Family \& Pets related) & 0.300$^{**}$ (0.148) \\ 
  topic 2 (Employment related) & $-$0.119 (0.184) \\ 
  topic 3 (Social media related) & 0.140 (0.165) \\ 
  topic 4 (Transportation related) & $-$0.172 (0.188) \\ 
  topic 5 (Food related) & 0.133 (0.194) \\ 
  topic 6 (COVID related) & $-$0.082 (0.200) \\ 
  topic 7 (Entertainment related) & $-$0.057 (0.161) \\ 
  topic 8 (People related) & 0.272 (0.198) \\ 
  comment is a question & 0.068 (0.049) \\ 
  length of parent comment & $-$0.003 (0.002) \\ 
  \textit{intercept} & 0.231$^{**}$ (0.115) \\ 
 \hline \\[-1.8ex] 
Observations & 8,369 \\ 
Log Likelihood & $-$14,899.820 \\ 
$\theta$ & 0.548$^{***}$  (0.013) \\ 
Akaike Inf. Crit. & 29,841.640 \\ 
\hline 
\hline \\[-1.8ex] 
\textit{Note:}  & \multicolumn{1}{r}{$^{*}$p$<$0.1; $^{**}$p$<$0.05; $^{***}$p$<$0.01} \\ 
\end{tabular} 
  \caption{Negative Binomial regression on score of the gif reply. The random-gif baseline is set as the reference category for models.} 
  \label{table:ScoreRegression} 
\end{table*}

\newcommand{\vbiggif}[1]{
\begin{minipage}{0.3\textwidth}
\href{http://giphy.com/gifs/#1}{
\includegraphics[width=0.3\linewidth]{gifs/#1.jpg} 
} \\\mbox{\small{#1}} \\
\end{minipage}
}

{
\begin{table*}[ht]
\centering
\begin{tabular}{p{0.30\textwidth} p{0.3\textwidth} p{0.3\textwidth}}
\toprule
Tag based & CLIP variant & \pepe \\
\midrule
\vbiggif{2gG2xiMTtFwsg} & \vbiggif {lfesfEtobCSbsHzC8d} & \vbiggif {tnYri4n2Frnig} \\
\vbiggif{fnjxvV295sWEJjvwXU} & \vbiggif {m9d3Xif3ShZ42CxlWP} & \vbiggif {5wWf7GR2nhgamhRnEuA} \\
\vbiggif{BAPSj0xM1cFe8} & \vbiggif {f9k1tV7HyORcngKF8v} & \vbiggif {5gw0VWGbgNm8w} \\
\vbiggif{iJsvRxNTAcup6DVfLP} & \vbiggif {loitbnzQ1JQ8Iizx8w} & \vbiggif {iXTrbbYMQBCMM} \\
\vbiggif{3oEjHLcg4QMU5umb9m} & \vbiggif {bfrlODgSLqXxS} & \vbiggif {65ODCwM00NVmEyLsX3} \\
\vbiggif{aKrTvuOv4hlKM} & \vbiggif {4HmjGg306HiLHWlm2f} & \vbiggif {26AHLBZUC1n53ozi8} \\
\vbiggif{3oKIPllDN24q8Awtwc} & \vbiggif {7J26CGAahos6d5S1A6} & \vbiggif {3o8doT9BL7dgtolp7O} \\
\vbiggif{jIu44mYwUItSHTW3tj} & \vbiggif {8hZ9FMolyKc0X8BSr7} & \vbiggif {Fq6Bdki3coEWQ} \\
\vbiggif{jTrWAzlFGfvVY34PSJ} & \vbiggif {iqkHA3DmB8GjORY030} & \vbiggif {3oEjHAUOqG3lSS0f1C} \\
\vbiggif{l396L17pwHWOIJrTG} & \vbiggif {OOzcnk3PzLDHqWs6Tb} & \vbiggif {KzyMcEfDh4Jiw} \\
\bottomrule
\end{tabular}
\caption{Examples of top 10 most frequently used gifs across all models in the RCT. Click an image to view the gif on Giphy. Images are ordered from most-used (top) to tenth-most (bottom).}
\label{tab:top10-gif-example}
\end{table*}
}

\begin{table*}[!htbp] \centering 
\begin{tabular}{@{\extracolsep{3pt}}lc} 
\\[-1.8ex]\hline 
\hline \\[-1.8ex] 
 & \multicolumn{1}{c}{\textit{Dependent variable:}} \\ 
\cline{2-2} 
\\[-1.8ex] & Cumulative number of replies received \\ 
\hline \\[-1.8ex] 
 gif reply score & 0.096$^{***}$ (0.010) \\ 
  post score & $-$0.0004$^{***}$ (0.0001) \\ 
  comment score & 0.0002 (0.0002) \\ 
  CLIP variant model & $-$0.196 (0.152) \\ 
  distribution-sampling model & $-$0.664$^{***}$ (0.160) \\ 
  \pepe model & $-$0.450$^{***}$ (0.138) \\ 
  Tag-based model & $-$0.195 (0.146) \\ 
  number of days after reply & $-$0.001 (0.001) \\ 
  comment text polarity & 0.048 (0.164) \\ 
  comment text subjectivity & $-$0.055 (0.147) \\ 
  topic 0 (Politics related) & $-$0.275 (0.430) \\ 
  topic 1 (Family \& Pets related) & $-$0.264 (0.412) \\ 
  topic 2 (Employment related) & $-$1.182$^{**}$ (0.549) \\ 
  topic 3 (Social media related) & 1.381$^{***}$ (0.421) \\ 
  topic 4 (Transportation related) & $-$0.021 (0.514) \\ 
  topic 5 (Food related) & $-$0.896 (0.567) \\ 
  topic 6 (COVID related) & $-$0.459 (0.564) \\ 
  topic 7 (Entertainment related) & $-$0.529 (0.452) \\ 
  topic 8 (People related) & $-$1.776$^{***}$ (0.647) \\ 
  comment is a question & 0.114 (0.133) \\ 
  length of parent comment & 0.0003 (0.007) \\ 
  \textit{intercept} & $-$1.877$^{***}$ (0.313) \\ 
 \hline \\[-1.8ex] 
Observations & 8,369 \\ 
Log Likelihood & $-$2,466.965 \\ 
$\theta$ & 0.143$^{***}$  (0.013) \\ 
Akaike Inf. Crit. & 4,977.930 \\ 
\hline 
\hline \\[-1.8ex] 
\textit{Note:}  & \multicolumn{1}{r}{$^{*}$p$<$0.1; $^{**}$p$<$0.05; $^{***}$p$<$0.01} \\ 
\end{tabular} 
  \caption{Negative Binomial regression on cumulative number of replies received. The random-gif baseline is set as the reference category for model comparison. } 
  \label{table:num-replies-regression} 
\end{table*}

\begin{table*}
\centering
\begin{tabular}{ccl}
\toprule \hline
\textbf{Topic} & \textbf{Dirichlet parameter} & \textbf{Keywords} \\
\midrule
0 & 0.1172 & people fuck trump shit make thing country n’t vote fucking \\ 
1 & 0.20164 & good time love kid make cat dog day year guy \\ 
2 & 0.09554 & pay work money people make job year buy time company \\ 
3 & 0.11245 & post make read people good time thing imgur video work \\ 
4 & 0.06541 & car live year drive day place time road city back \\ 
5 & 0.05672 & eat make food good water drink taste cheese pizza coffee \\ 
6 & 0.06662 & people covid die vaccine life make work problem mask n’t \\ 
7 & 0.0888 & movie play game good watch show love great time song \\ 
8 & 0.02752 & wear mask red shirt woman hair white man hat black \\ 
9 & 0.14292 & back make put hand time guy car head thing big \\
\hline
\bottomrule
\end{tabular}
\caption{Topic modeling keywords for Imgur Comments} 
\label{table:topic-model-keywords} 
\end{table*}

\section{GIF categories on GIPHY}
\label{appendix:GIFCategories}

\begin{supertabular}{l p{0.25\textwidth}}
\toprule
\textbf{Category} &               \textbf{Subcategory} \\
\midrule
        Reactions &                               what \\
        Reactions &                          hair flip \\
        Reactions &                              bored \\
        Reactions &                              frown \\
        Reactions &                          slow clap \\
        Reactions &                           mic drop \\
        Reactions &                            goodbye \\
\bottomrule
\end{supertabular}

\section{List of selected tags from GIPHY}
\label{appendix:SelectedTags}
adorable, agreed, amazing, amused, angry, annoyed, anxiety, anxious, applause, approval, approve, aw, awesome, awkward, bad, beautiful, best wishes, blank stare, blink, blush, bored, bow, bravo, but why, buy, bye, captivated, celebrate, cheeky, cheering, cheers, clap, come on, comic, compliment, compliments, concerned, confused, congratulations, cool, crazy, creeping, cringe, crushing, cry, curtsy, cute, damn, dance, dancing, deadpan stare, debate, depressed, dickhead, disagree, disappointed, disapprove, disbelief, disgust, dislike, diss, divertente, dont care, doubt, doubtful, drink, drinking, drunk, dubious, dying, eating, eating popcorn, embarassed, engrossed, ennui, excited, face palm, faint, fingers crossed, flirt, flushed, freaking out, frustrated, fuck, fun, funny, gagging, get well, glare, good luck, gossip, grateful, gratitude, great, great job, grin, hahahah, happy, happy dance, head shake, hide, high five, hilarious, honestly, hope, horror, hugs, hugs love, hysterical, ill, impressed, incredulous, insult, interested, interesting, judge, judging you, just, keep going, kiss, laugh, leaving, lets go, lies, like, looking, looking around, love you, lovely, luv u, luv you, mad, mind blown, mock, motivational, moved, muah, much appreciated, nah, nasty, need, nervous, nice one, no, nod, not amused, not funny, not interested, oh shit, overwhelmed, panic, partying, perfect, pissed, please, pleased, pointing, praise, pray, pregnant, proud, pumped, questioning, raises hand, realization, relief, respect, reunited, roast, roll eyes, sad, sadness, salute, sarcastic, savage, scared, scary, screaming, secret, seriously, sexy, shame, shock, shook, shrug, shut up, shy, sigh, sips tea, sitting, sleepy, sloth, smart, smile, smug, sobbing, sorpren, sorry, spit, stoked, stressed, stunned, success, sudden realization, surprise, suspicious, sweating, swoon, swooning, take notes, tantrum, tears, thank, think, thirsty, thumbs down, thumbs up, tired, too funny, touched, unamused, unbelievable, uncomfortable, unhappy, unimpressed, unsure, upset, vomit, waiting, wave, weary, weird, whatever, will, wince, wink, wrestling, yawn, yell, yes, yum

\section{List of filtering keywords on Imgur experiment}
\label{appendix:ImgurFilterList}
depression, depressing, mental, health, death, dead, alcohol, alcoholism, weed, drugs, addiction, covid, beer, stoned, black, white, arabic, hispanic, latino, latina, latinx, police, cop, racism, racists, race, sexism, sexist, sexy, armed, overthrow, government, republican, democrats, maga, liberal, liberals, conservative, conservatives, offender, victim, disability, disabled, jerking, PD, gun, shots, fired, cops, officer, officers, killing, murder, murdered, kill, kills, killed, murders, shoot, taser, bystander, trigger, handgun, pansexual, sexuality, homosexual, gay, lesbian, corona, virus, coronavirus, vaccine, vaccinated, viruses, vaccination, die, fascist, fascists, antifa, sharia, islam, islamic, christian, jewish, muslim, blasphemy, blasphemic, death, conviction, church, priest, pastor, religious, religion, sharia, shia, sunni, judge, bible, qaran, torah, hindu, hindus, christians, jew, jews, muslims, islamist, execute, murder, captive, captives, malpractice, insurance, insured, threat, threatening, war, troops, violence, fighting, conflict, medicine, prescription, drug, dying, hospice, life, doctor, hospital, nurse, pedophiles, pedophile, bitch, republicans, democrat, coup, tax, recession, pedo, criminal, criminals, politician, politicians, health, healthcare, america, american, voter, voting, votes, vote, voters, citizen, immigrants, immigrant, citizens, candian, canada, eu, european, trump, red, blue, cancer, slavery, slaves, slave, disease, sickness, sorry, nazi, nazis, death, pro-death, pro-life, profile, abortion, aborted, aborting, victims, jail, whore, slut, rape, raped, raping, behead, beheadings, beheaded, torture, tortured, torturing, taliban, afghanistan, soldier, soldiers, kabul

\end{document}